\documentclass[letterpaper, 10 pt, conference]{ieeeconf}
\IEEEoverridecommandlockouts
\usepackage{graphics} 
\usepackage{epsfig} 
\usepackage{mathptmx} 
\usepackage{times} 
\usepackage{amsmath} 
\usepackage{amssymb}  
\usepackage{siunitx} 
\usepackage{textcomp}
\usepackage{gensymb} 
\usepackage[colorlinks=true]{hyperref}
\hypersetup{
  colorlinks=true,
  linkcolor=blue,
  urlcolor=cyan,
}
\usepackage{booktabs}
\usepackage{multirow}
\usepackage{xcolor}
\usepackage{booktabs}
\usepackage{url}
\usepackage{xspace}
\usepackage{booktabs}
\usepackage{soul}
\usepackage{makecell}
\usepackage{multicol}
\usepackage{rotating}
\usepackage[percent]{overpic}
\usepackage[verbose=silent]{microtype}
\usepackage{contour}
\usepackage{courier}
\usepackage{algpseudocode}
\usepackage{algorithm}
\usepackage{wrapfig}
\usepackage{float}
\usepackage[numbers,sort&compress]{natbib}
\usepackage[moderate]{savetrees}

\newcommand{\CAIR}{
1
}
\newcommand{\TRI}{
2
}

\newcommand{\mypara}[1]{\textbf{{#1}}}

\newcommand{\ie}{\textit{i}.\textit{e}. }
\newcommand{\eg}{\textit{e}.\textit{g}. }

\newcommand{\rewardBaseline}{R_{\textrm{Unf}}}
\newcommand{\rewardOurs}{R_{\textrm{CA}}}
\newcommand{\rewardCanon}{R_{\textrm{C}}}
\newcommand{\rewardAlign}{R_{\textrm{A}}}

\title{\LARGE \bf 
Cloth Funnels: Canonicalized-Alignment\\
for Multi-Purpose Garment Manipulation
}

\author{Alper Canberk\textsuperscript{\CAIR}, Cheng Chi\textsuperscript{\CAIR}, Huy Ha\textsuperscript{\CAIR}, \\Benjamin Burchfiel\textsuperscript{\TRI}, Eric Cousineau\textsuperscript{\TRI}, Siyuan Feng\textsuperscript{\TRI} and Shuran Song\textsuperscript{\CAIR} \\
\href{https://clothfunnels.cs.columbia.edu}{clothfunnels.cs.columbia.edu}
\thanks{\textsuperscript{\CAIR}  Columbia University}%
\thanks{\textsuperscript{\TRI} Toyota Research Institute}%
}
\begin{document}
\maketitle

\begin{abstract}
Automating garment manipulation is challenging due to extremely high variability in object configurations. 
To reduce this intrinsic variation, we introduce the task of ``canonicalized-alignment'' that simplifies downstream applications by reducing the possible garment configurations. This task can be considered as ``cloth state funnel'' that manipulates arbitrarily configured clothing items into a predefined deformable configuration (i.e. canonicalization) at an appropriate rigid pose (i.e. alignment). In the end, the cloth items will result in a compact set of structured and highly visible configurations -- which are desirable for downstream manipulation skills. 
To enable this task, we propose a novel canonicalized-alignment objective that effectively guides learning to avoid adverse local minima during learning. Using this objective, we learn a multi-arm, multi-primitive policy that strategically chooses between dynamic flings and quasi-static pick and place actions to achieve efficient canonicalized-alignment.
We evaluate this approach on a real-world ironing and folding system that relies on this learned policy as the common first step.
Empirically, we demonstrate that our task-agnostic canonicalized-alignment can enable even simple manually-designed policies to work well where they were previously inadequate, thus bridging the gap between automated non-deformable manufacturing and deformable manipulation.
\end{abstract}

\section{Introduction}

Why has garment manipulation proved more difficult to automate than more typical rigid and articulated objects?
We argue that two key factors are severe self occlusion, which is present in the large set of possible crumpled states, and the infinite degrees of freedom inherent to clothing. 
As a result, it is impractical to manually define manipulation policies that achieve reliable manipulation — a cornerstone of current automated non-deformable manufacturing pipelines.
In this work, we explore bridging the gap between existing approaches to automation and the challenging domain of clothing.
We show that when a robot first manipulates arbitrarily configured clothing items into a predefined configuration (\ie canonicalization) at an appropriate pose (\ie alignment), downstream manipulation skills work significantly more reliably.

\begin{figure}
    \centering
    \includegraphics[width=0.99\linewidth]{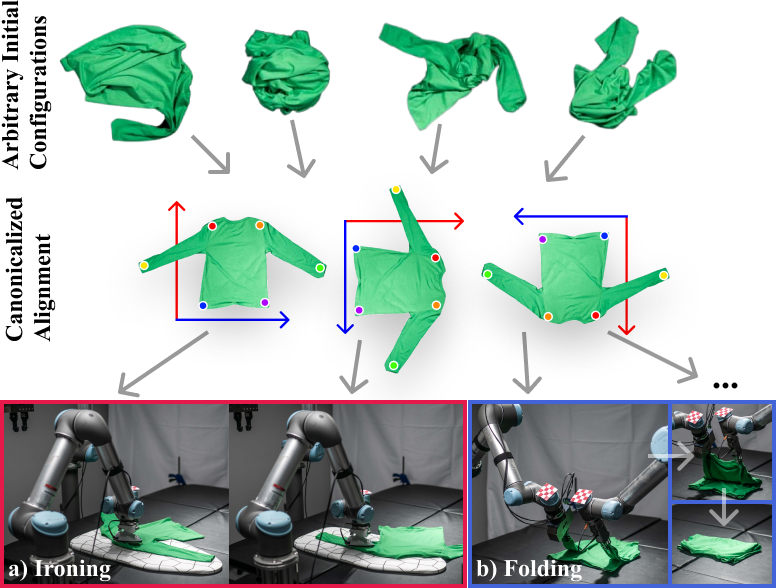} \vspace{-6mm}
    \caption{Canonicalized-Alignment \emph{funnels} the large space of possible cloth configurations into a much smaller and better structured set of highly-visible states that greatly simplifies downstream tasks such as ironing or folding.} 
    \vspace{-5mm}
    \label{fig:teaser}
\end{figure}

\begin{figure*}
    \centering
    \includegraphics[width=\linewidth]{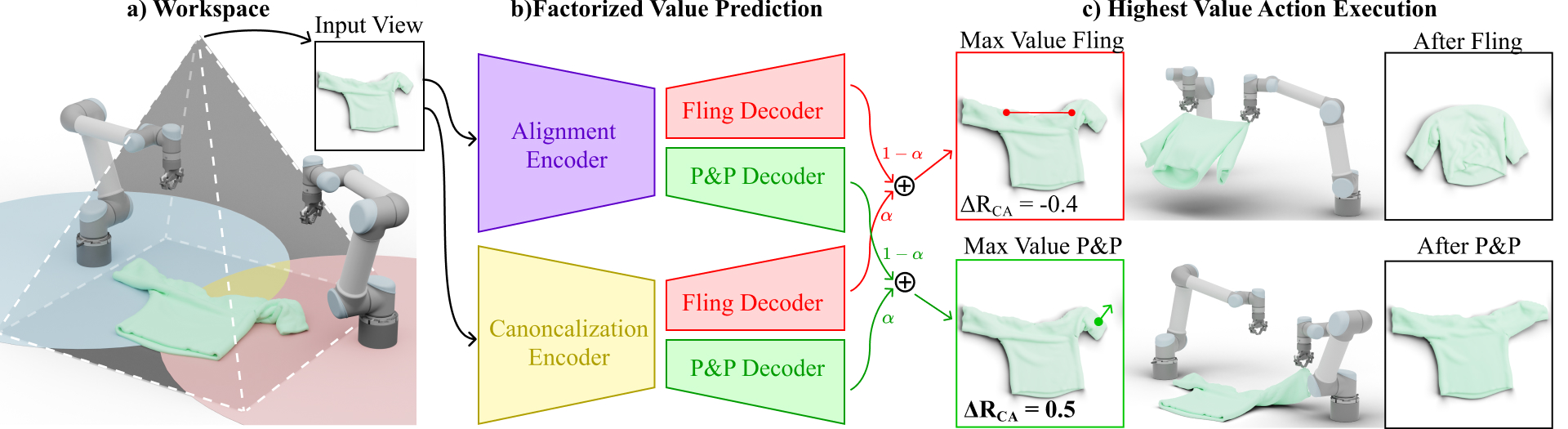} 
    \vspace{-7mm}
    \caption{\textbf{Approach Overview.} a) A batch of scaled and rotated observations are created from a top-down RGB image of the workspace and then concatenated with a scale-invariant coordinate map. b) The batch of inputs is fed through the factorized network architecture, producing a batch of rotated and scaled value maps for each primitive. c) All primitive batches are concatenated and the maximum value pixel parameterizes the action to be executed.}
    \vspace{-7mm}
    \label{fig:pipeline}
\end{figure*}

Recently, real-world cloth manipulation has received significant attention.
Some of the earliest cloth manipulation work explored manually designed heuristics which worked well for specific clothing types, configurations, and tasks, such as cloth unfolding~\cite{cusumano2011bringing,maitin2010cloth,triantafyllou2016geometric,osawa2007unfolding}, smoothing~\cite{Sun2013AHA,willimon2011model}, folding~\cite{zhou2021folding,maitin2010cloth,osawa2006clothes,miller2011parametrized}, 
but their strong assumptions initial states, fiducial markers, specialized tools, or cloth type/shape do not generalize.
%
More recently, learning-based approaches have shown success in more general cloth manipulation behavior. 
One line of work has explored supervised-learning from human demonstrations for smoothing~\cite{ken2022speedfolding} and folding~\cite{zhou2021folding}, but those methods required costly human demonstrations/annotations. 
Another recent line of work employs fully self-supervised learning and has shown success in learning to unfold~\cite{ha2021flingbot} (but doesn't generalize to other tasks) and in tackling visual goal-conditioned manipulation of a single square cloth instance~\cite{lee2020learning}.

Instead of learning arbitrary monolithic cloth manipulation tasks, we hypothesize that it is more efficient to learn a robust task-agnostic canonicalization and alignment policy from which other task-specific manipulation skills may be chained.
This is because such a policy \emph{funnels} unstructured and self-occluded cloth configurations into structured states with clearly visible key points (Fig.~\ref{fig:teaser}, middle), reducing the complexity of the task-specific downstream policy, and enabling even simple heuristics to work with a high success rate.

To this end, we define a new ``canonicalized-alignment'' task for garment manipulation, where the goal is to transform a garment from its arbitrary initial state into a canonical shape (defined by its category) and align it with a particular 2D translation and rotation.
The end result is a decomposition of garment manipulation into two factorized parts.
The first part funnels a diverse set of cloth configurations into SE-2 transforms of a small set of states with high visibility. 
The second part consists of downstream behaviors which relies on kinematically feasible transforms of structured initial configurations and full keypoint observability to achieve high task success rates.

Our primary contribution is the introduction of ``canonicalized-alignment'', a garment manipulation task which serves as a cloth funnel for reducing general-purpose garment manipulation complexity.
We achieve this by the following technical contributions: 
\begin{itemize}
    \item  We propose a learned multi-arm, multi-primitive manipulation policy that strategically chooses between dynamic flings and quasi-static pick\&place actions to efficiently and precisely transform the garment into its canonicalized and aligned configuration.
    
    \item To train the policy, we proposed a novel factorized reward function that avoids adverse local minima which plague the generic goal-reaching formulations by decoupling deformable shape and rigid pose.
    
    \item We evaluate our approach in multiple downstream garment manipulation tasks in the real-world on a physical robot, including folding and ironing.
\end{itemize}
Our experiments show that incorporation of canonicalized-alignment significantly reduces the complexity of downstream applications, suggesting that robust canonicalized-alignment provides a practical step forward toward multi-purpose garment manipulation from arbitrary states for diverse tasks. 

\section{Related Work}

\textbf{Heuristic-based cloth manipulation.}
Heuristic-based manipulation pipelines -- where action selection and planning is manually designed -- can produce impressive results.
However, the generality and robustness of these approaches is limited due to strong assumptions regarding pre-canonicalized initial state~\cite{li2019learning,miller2011parametrized}, fiducial markers~\cite{bersch2011bimanual}, specialized tools~\cite{osawa2006clothes}, and cloth type and shape~\citep{Sun2013AHA,doumanoglou2016folding,maitin2010cloth,tanaka2007origami,balkcom2008robotic,willimon2011model,cusumano2011bringing,stria2014garment,osawa2007unfolding}.



\textbf{Learning-based cloth unfolding.}
Learning-based methods can self-discover the best policies for a distribution of cloths using real-world self-supervision~\cite{ha2021flingbot,xu2022dextairity} or simulator states~\cite{huang2022mesh,lin2022learning}. 
While these approaches have been successfully applied to cloth unfolding~\cite{ha2021flingbot,xu2022dextairity} or canonicalization~\cite{huang2022mesh}, they do not consider canonicalized-alignment.
This limits their applicability since heuristic-based pipelines cope poorly with unmet cloth assumptions or kinematic constraints.


\textbf{Goal-conditioned cloth manipulation.}
Towards generic goal-conditioned cloth manipulation, prior works have investigated reinforcement learning~\cite{matas2018sim,hietala2021closing,jangir2020dynamic,tsurumine2019deep}, real-world self-supervised learning~\cite{lee2020learning} and imitation learning~\cite{seita2020deep}.
However, these methods often struggle to bridge the sim2real gap~\cite{jangir2020dynamic}, generalize across cloth instances~\cite{lee2020learning,weng2022fabricflownet,seita2020deep} or generalize between garment types~\cite{tanaka2018emd, hietala2021closing, tsurumine2019deep}.
Furthermore, all goal-conditioned works do not address how goal vertices/key points/images can be obtained for a completely novel cloth instance.
Instead, our proposed approach can accommodate different garment categories and generalize to a variety of novel real-world garment instances from simulation training.




\section{Method}
\label{sec:method}

\subsection{A Multi-Purpose Garment Manipulation Pipeline}
\label{sec:method:recipe}

We propose a factorized approach to multi-purpose garment manipulation from arbitrary states that decomposes the process into two steps.
First, the robot executes a learned \emph{task-agnostic} canonicalized-alignment policy, which leaves the garment in a known configuration predefined for the clothing category at a specified 2D rotation and translation.
Second, the robot executes a \emph{task-specific} keypoint based policy, which could be as simple as a manually-designed heuristic.
This approach confers three primary benefits:
\begin{itemize}
    \item \textbf{Arbitrary initial configuration}:
    Canonicalization \emph{funnels} the large space of possible cloth configurations into a narrow distribution of highly-structured fully-observable configurations from which downstream policies can more easily operate.
    \item \textbf{Downstream task-awareness}:
    Flexible goal-conditioned alignment allows the canonicalized cloths to be placed at specified positions and orientations that are kinematically appropriate for particular downstream tasks.
    \item \textbf{Clothing category generalization}:
    A keypoint-based cloth representation effectively reduces the observation space from having to represent the infinite DoF down to a few meaningful keypoints.
    Further, cloths are always in a known canonicalized configuration.
    These two properties combined not only simplifies learning downstream task-specific manipulation policies, but also makes it possible to engineer heuristics that work reliably for a clothing category.
\end{itemize}
Next, we will discuss how the canonicalized-alignment task is formulated (Sec.~\ref{sec:method:can_align_task}), learned (Sec.~\ref{sec:spatial_policy}), and implemented alongside the several downstream task policies (Sec.~\ref{sec:method:heuristics}).

\subsection{The Canonicalization \& Alignment Task}
\label{sec:method:can_align_task}

\begin{figure}
    \centering
    \includegraphics[width=\linewidth]{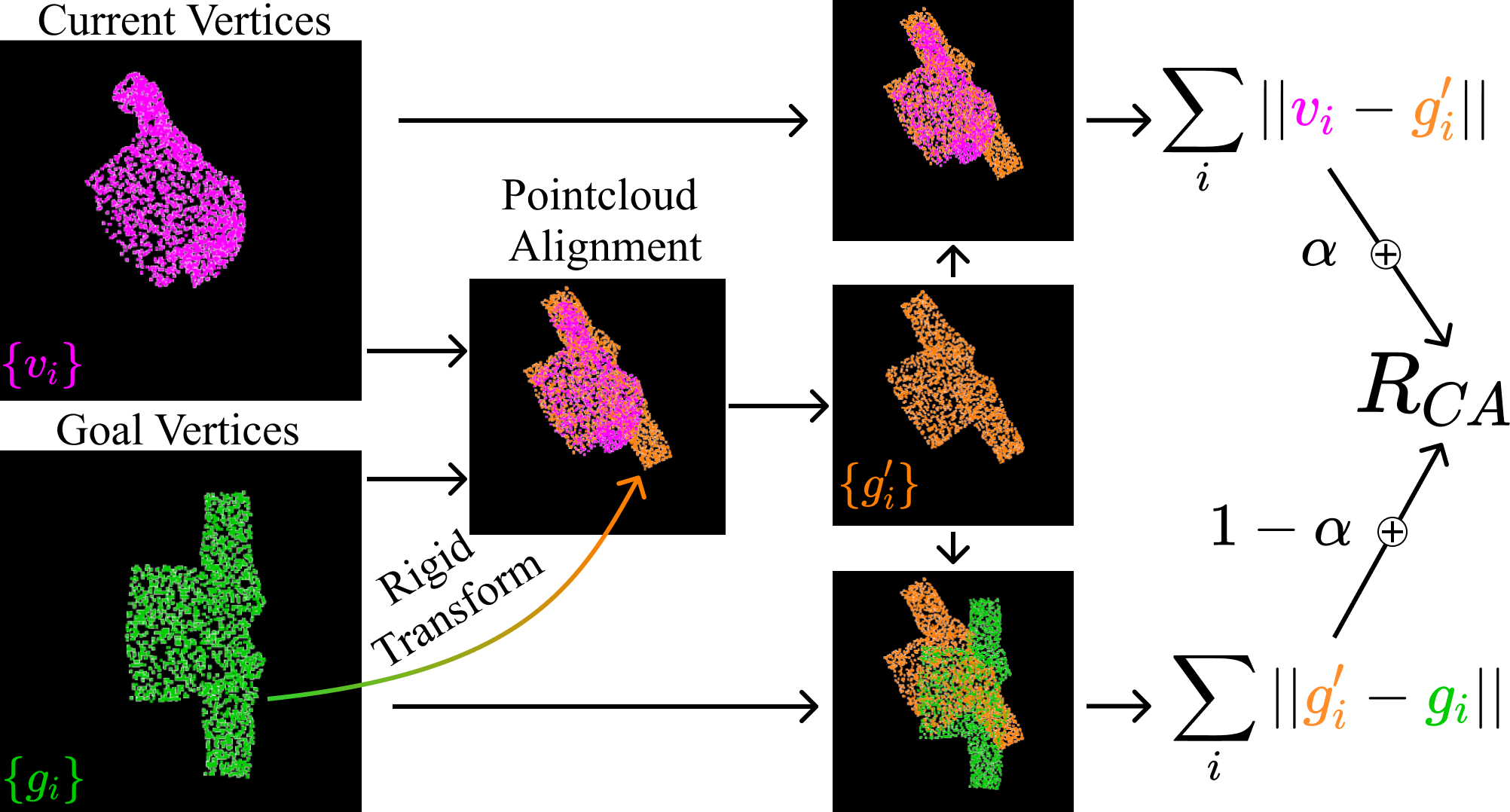}
    \vspace{-6mm}
    \caption{\textbf{Reward Computation.}
    From the goal configuration $\mathbf{g}$ (green) and current configuration $\mathbf{v}$ (magenta), we compute a best-alignment configuration $\mathbf{g}^\prime$ (orange).
    Then, the average vertex distance between $\mathbf{g}^\prime$ and $\mathbf{g}$ (where only rigid transforms matter) gives the alignment reward $\rewardAlign$, while that between $\mathbf{g}$ and $\mathbf{v}$ (where only deformation matters) gives the canonicalization reward $\rewardCanon$.
    Our factorized reward $\rewardOurs$ is a weighed sum between $\rewardAlign$ and  $\rewardCanon$.
    }
    \vspace{-6mm}
    \label{fig:reward_formulation}
\end{figure}

\mypara{Problem Formulation.}
Given a clothing item in some clothing category in an arbitrary initial configuration, the goal of canonicalization is to reach the human-defined standard deformable configuration for that clothing category, such as a T-shaped configuration for shirts and the upside-down V-shaped configuration for pants.
Note that this only accounts for the garment's shape, but not its pose in the workspace.
This means canonicalization alone can't ensure that downstream tasks are kinematically feasible.
To address this shortcoming, the goal of ``canonicalized-alignment'' is to reach canonicalization at a specific planar position and rotation in the workspace.

\mypara{Naive Reward Formulation.}
Given a simulated cloth instance with $N$ vertices, let  $\mathbf{v} = \{v_{i}\}_{i\in[N]}$ denote the current configuration of the cloth (Fig.~\ref{fig:reward_formulation}, magenta), where each $v_i \in \mathbb{R}^3$ is the position of the $i$th cloth vertex.
Given a goal configuration $\mathbf{g} = \{g_{i}\}_{i\in[N]}$, the average per-vertex distance between $\mathbf{g}$ and $\mathbf{v}$ gives a generic goal-conditioned cloth manipulation cost.
In the specific case where $\mathbf{g}$ is a canonicalized configuration of the cloth at the goal position and rotation (Fig.~\ref{fig:reward_formulation}, green), we have the following straightforward canonicalized-alignment reward 
\begin{equation}
    \rewardBaseline  = -\lvert\lvert \mathbf{g} - \mathbf{v} \rvert\rvert_{2}
    \label{eq:unfactorized}
\end{equation}
Clearly,  $\rewardBaseline$ is consistent, in that a policy which achieves $\rewardBaseline=0$ achieves perfect canonicalized-alignment.
However, this formulation has two primary downsides:
\begin{enumerate}
    \item \textbf{Entangled supervision.}
    When $\rewardBaseline$ is low, it can be difficult for the policy to tell whether it should make a planar transformation of the cloth configuration (such as shifting entire cloth to the right) or a deformable adjustment (such as flipping a shirt's sleeve outwards).
    \item \textbf{Over-emphasis on cloth pose.}
    Actions that shift the cloth result in sharp and large changes in $\rewardBaseline$, while actions of smaller magnitudes become insignificant.
    Since such small adjustment actions are required to bring a poorly canonicalized cloth to a well-canonicalized one,  $\rewardBaseline$ fails to put enough emphasis on the canonicalization subtask, and leads to a problematic local minima in policy learning.
\end{enumerate}

\mypara{Factorized Reward Formulation.}
To alleviate these shortcomings, we propose a reward factorization, that expresses the canonicalization $\rewardCanon$ and alignment $\rewardAlign$ aspects of the task separately:
\begin{align} 
    \label{eq:factorized}
    \rewardOurs & = (1-\alpha) \rewardCanon + \alpha \rewardAlign
\end{align}
where $\alpha \in (0,1)$ is a hyperparameter.
With this factorization, we can provide separate supervision $\rewardCanon$ and $\rewardAlign$ during training, while acting with respect to $\rewardOurs$ during data-collection and inference.
This helps the policy distinguish how actions separately affect the cloth shape or planar pose. 
Further, with a tunable $\alpha$, we can emphasize $\rewardCanon$ more than $\rewardAlign$, which leads to a better canonicalization.

To factorize the reward, we propose to compute a transform $T$, which transforms $\mathbf{g}$ into a best-aligned goal configuration $\mathbf{g}^\prime$ (Fig.~\ref{fig:reward_formulation}, orange).
Given such a $\mathbf{g}^\prime$, its distance to $\mathbf{v}$ accounts only for their deformable shape mismatch, which serves as the canonicalization reward,
\begin{equation}
    \rewardCanon = 
    - \lvert\lvert \mathbf{v} - \mathbf{g}^\prime\rvert\rvert_2
    \label{eq:rigid}
\end{equation}
Meanwhile, by $T$'s definition, the distance between $\mathbf{g}^\prime$ and $ \mathbf{g}$ accounts only for the mismatch in planar position and rotation, which serves as the alignment reward,
\begin{equation}
    \rewardAlign = 
    -\lvert\lvert \mathbf{g}^\prime - \mathbf{g}\rvert\rvert_2.
    \label{eq:deform}
\end{equation}

\mypara{Factorization Implementation.}
To find $\mathbf{g}^\prime$, we have observed that naively minimizing the average per-vertex distance between $\mathbf{g}$ and $\mathbf{v}$ is extremely sensitive to outliers, so does not give us the best alignment.
Such outliers arise due to mismatches in $\mathbf{g}$'s and $\mathbf{v}$'s deformable shapes where small protrusions with large offsets (\eg a shirt's arm folded inwards) could significantly shift the minimum distance configuration.
To filter out such outliers, we optimize the transform $T$ which minimizes this distance for only a subset of points, where point $i$ is included if $\lvert\lvert g_i - v_i\rvert\rvert_2 \leq \tau$ for some scalar threshold $\tau$ then apply $T$ to $\mathbf{g}$ to get $\mathbf{g}^\prime$.
We repeat this minimization and filter procedure in iterations, using the previous iteration's $\mathbf{g}^\prime$ as the current $\mathbf{g}$, until convergence.

In our experiments, we observed that $\alpha = 0.6$ and $\tau=0.3$ performs best.
To account for different cloth sizes, we normalize all $\rewardCanon$, $\rewardAlign$, $\rewardBaseline$, and $\tau$ by the geometric mean of the cloth's height and width in a canonicalized configuration.
Since most garments are mirror-symmetric, we select the highest reward from either the goal configuration or its mirror-flip in the goal's local frame.

\subsection{Multi-Primitive Spatial Action Maps Policy}
\label{sec:spatial_policy}

Coarse-grain dynamic multi-arm flings can efficiently unfold and align garments from crumpled states~\cite{ha2021flingbot}, but are insufficient for the fine-grained adjustments required to achieve canonicalization.
To overcome this challenge, we propose a multi-arm, multi-primitive system that combines quasi-static and dynamic actions, which enables both efficient \emph{and} fine-grained manipulation. To unify the primitive parametrizations and easily enforce constraints, we use a spatial action maps policy.

\mypara{Spatial Action Maps} is a convolutional neural network~\cite{lecun2010convolutional} (CNN) policy for learning value maps~\cite{Wu_2020} where actions are defined on a pixel grid.
Through its simple and effective exploitation of translational, rotational, and scale equivariances, spatial action maps is a popular framework for learning robotic policies~\cite{wu2020spatial,ha2021flingbot,xu2022dextairity}.

We extend FlingBot~\cite{ha2021flingbot}'s spatial action maps approach as follows.
Given a $H\times W \times 3$ top-down view of the workspace (Fig.~\ref{fig:pipeline}a), we rotate and scale it to form a stack of transformed observations of shape $K \times H\times W \times 3$.
To help the network reason about the cloth's alignment, we concatenate a $K \times H \times W \times 2$ scale-invariant, normalized (between -1 and 1) positional encoding to the transformed observation stack.
To enable multiple primitives, we propose a factorized network architecture (Fig.~\ref{fig:pipeline}b) with two encoders, one for each task's reward ($\rewardCanon$, $\rewardAlign$), where each encoder has two decoder heads, one for each primitive.
The encoders take in the transformed observation stack, and the decoders output value maps, one for each reward-primitive pair.
The value maps are combined using \eqref{eq:factorized}, and the highest value action (over all action parameters and primitives) is chosen (Fig.~\ref{fig:pipeline}c).

\mypara{System Implementation.} 
In our experiments, we consider two primitives, quasi-static pick\&place and dynamic flings.
We use $(H,W) = (128, 128)$
and a decaying $\epsilon$-greedy with $\epsilon = 1$ for exploration of
1) action primitives (\ie fling v.s. pick\&place) with half life 5000, and 
2) action parameters within each primitive with half life 2500.
By constraining the observation's transforms to 16 rotations spanning $360^\circ$ and scales in $\{0.75, 1.0, 1.5, 2.0, 2.5, 3.0\}$ (giving $K=96$), we can ensure that arms neither collide nor cross-over each other. 
Our value networks' predictions are supervised using the delta-reward values -- which is the difference in $\rewardOurs$ before and after an action is taken -- using the MSE loss and the Adam~\cite{kingma2014adam} optimizer with a learning rate of $1e-4$.
We train our model for 12,500 episodes which takes 2 days on 4 NVIDIA RTX3090s.

\begin{figure}
    \centering
    \includegraphics[width=0.98\linewidth]{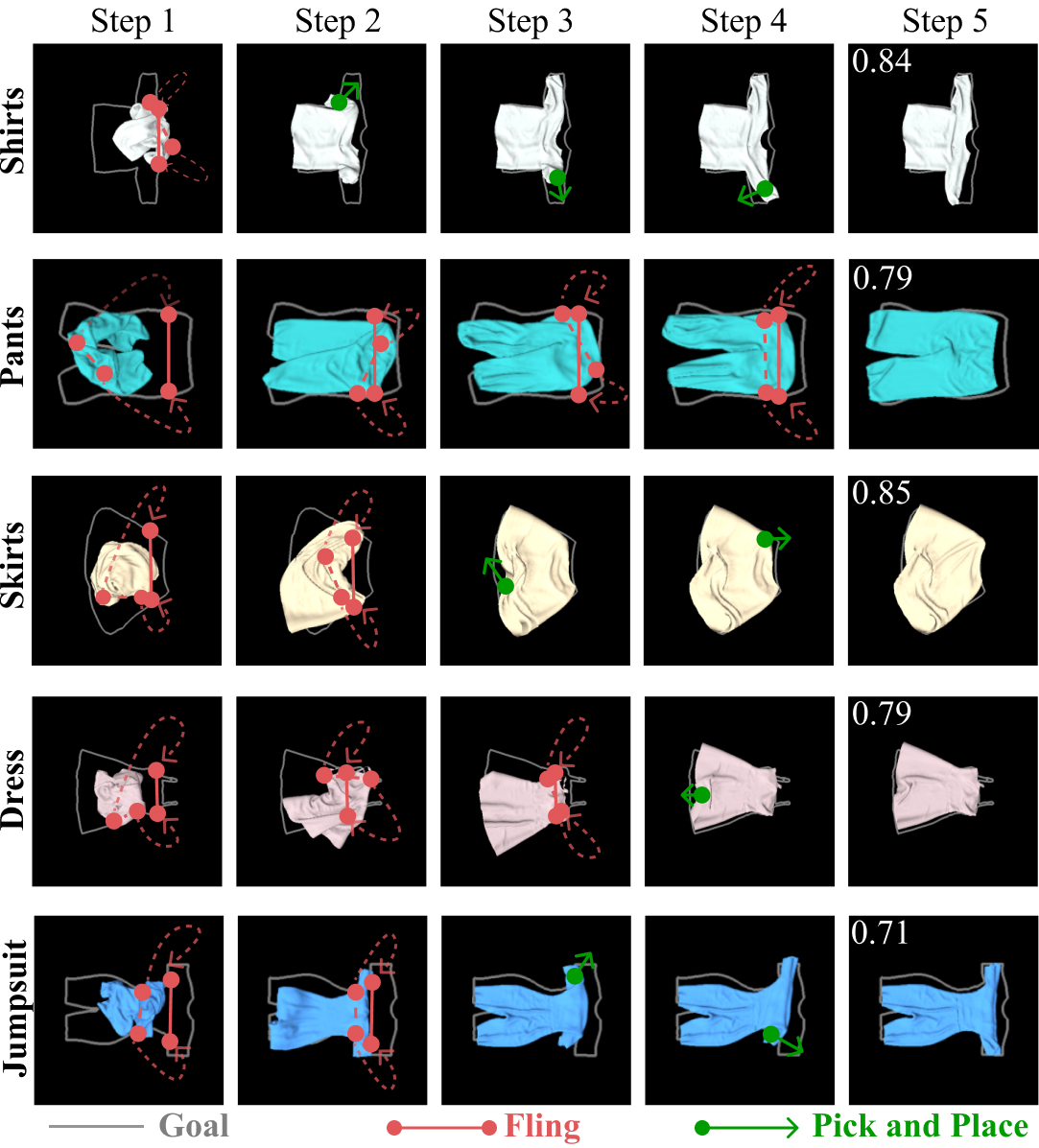}
    \vspace{-3mm}
    \caption{
    \textbf{Canonicalized-Alignment of Multiple Categories.}
    In each row, we demonstrate a sequence of 5 actions taken by the model corresponding to a clothing category in simulation.}
    \vspace{-6mm}
    \label{fig:canal:sim}
\end{figure}

\subsection{Keypoint-based Task Heuristics}
\label{sec:method:heuristics}

Compared to learning-based approaches, heuristics are highly interpretable and thus simple to define.
Here, we demonstrate that it's possible to use heuristics for shirt ironing with no keypoints and folding with a small set of keypoints.

\mypara{Keypoint Detection.}
We collect 200 cloth configurations from simulation with coverage at least $60\%$ and trained a DeepLabv3~\cite{chen2019rethinking} detector for each garment class.
Using a random 80/20 training/evaluation split, we observed that this detector generalizes well to novel garment instances with average error of 5/128 pixels.
Setting up a keypoint detector model for a new clothing category takes roughly 1 hour.
After detection, each keypoint is depth-projected into 3D points and transformed into the workspace frame of reference.
By representing cloths as a set keypoints, we sidestep  their infinite DoF by using a few meaningful keypoints as the representation, which makes it simpler to define heuristics over them.
For instance, long sleeve shirts have six keypoints for two sleeves, shoulders, and waists.

\begin{figure}
    \centering
    \includegraphics[width=0.98\linewidth]{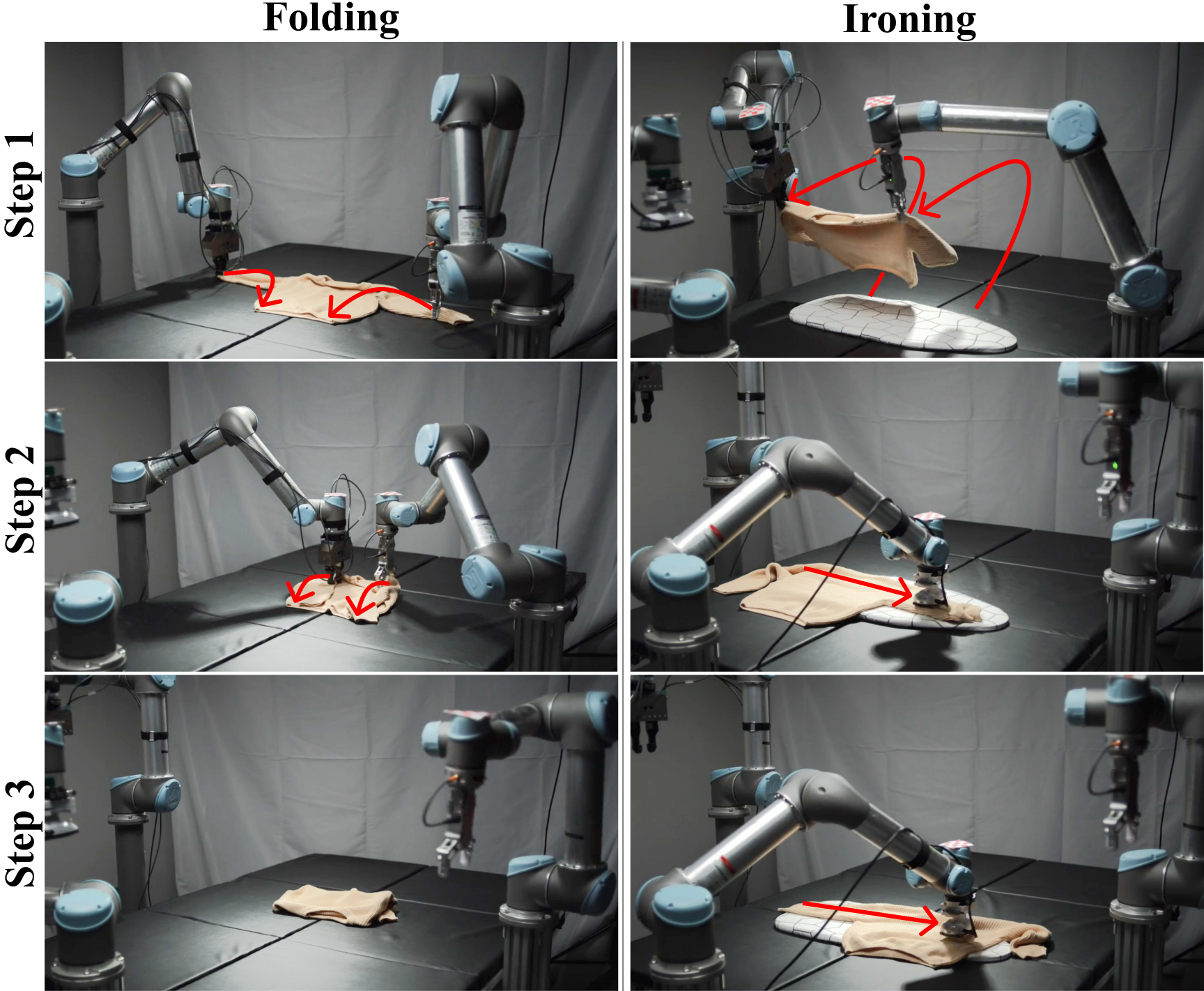} \vspace{-3mm}
    \caption{
        \textbf{Real-world Ironing \& Folding.}
        Reliable canonicalized-alignment not only gives high-visibility starting configurations, which reduces complexity for downstream tasks like folding (step 1), but can also be called multiple times with different target alignments, which is useful in ironing (step 2 \& 3).
        More video results on the \href{https://clothfunnels.cs.columbia.edu}{project website}.
    }
    \label{fig:downstream:real}  \vspace{-3mm}
\end{figure}

\mypara{Ironing Heuristic.}
For garment manipulation pipelines, specialized tools are placed at a fixed location in the workspace.
For ironing, the extra tools involved are the ironing board and the arm holding the iron (Fig.~\ref{fig:downstream:real} left).
Given a well canonicalized and aligned shirt, an open-loop ironing primitive where the end effector moves from one end of the ironing board and back without any perception can be sufficient.
In our setup, we use two transforms such that the left and right side of the shirt is on the ironing board respectively.

\mypara{Folding Heuristic.}
First, the sleeves are folded towards the waist using a dual-arm pick and place action.
Here, the pick point is the sleeve keypoint, while the place point is the quarter and three-quarter point along the waist line (computed from the waist keypoints).
Since not all shirt arms are long enough to reach the waist points, the place points are constrained to be an arm's length distance away from the shoulder keypoints.
The arm length can be computed from keypoints as the minimum distance between the sleeve and the shoulder keypoints over the left and right arms.
In the second step, with the arms folded in, the shoulder keypoints are picked and placed at the waist keypoints (Fig.~\ref{fig:downstream:real} right).
\section{Results}

\textbf{In simulation}, we conduct ablation studies of reward formulation (Sec.~\ref{sec:eval:reward_formulation}) and action primitives (Sec.~\ref{sec:eval:combined_primitives}).
Next, we demonstrated our approach on five garment categories from Cloth3D~\cite{bertiche2020cloth3d} (Sec.~\ref{sec:eval:multicategory}) and a folding downstream task (Sec.~\ref{sec:eval:downstream}).
\textbf{In the real world}, we include primitive and reward comparisons for the long-sleeve shirt category on canonicalized-alignment (Tab.~\ref{tab:primitive:real}, Fig.~\ref{fig:canal:real}), folding and ironing. 


\begin{figure}
    \centering
    \vspace{-2mm}
    \includegraphics[width=0.98\linewidth]{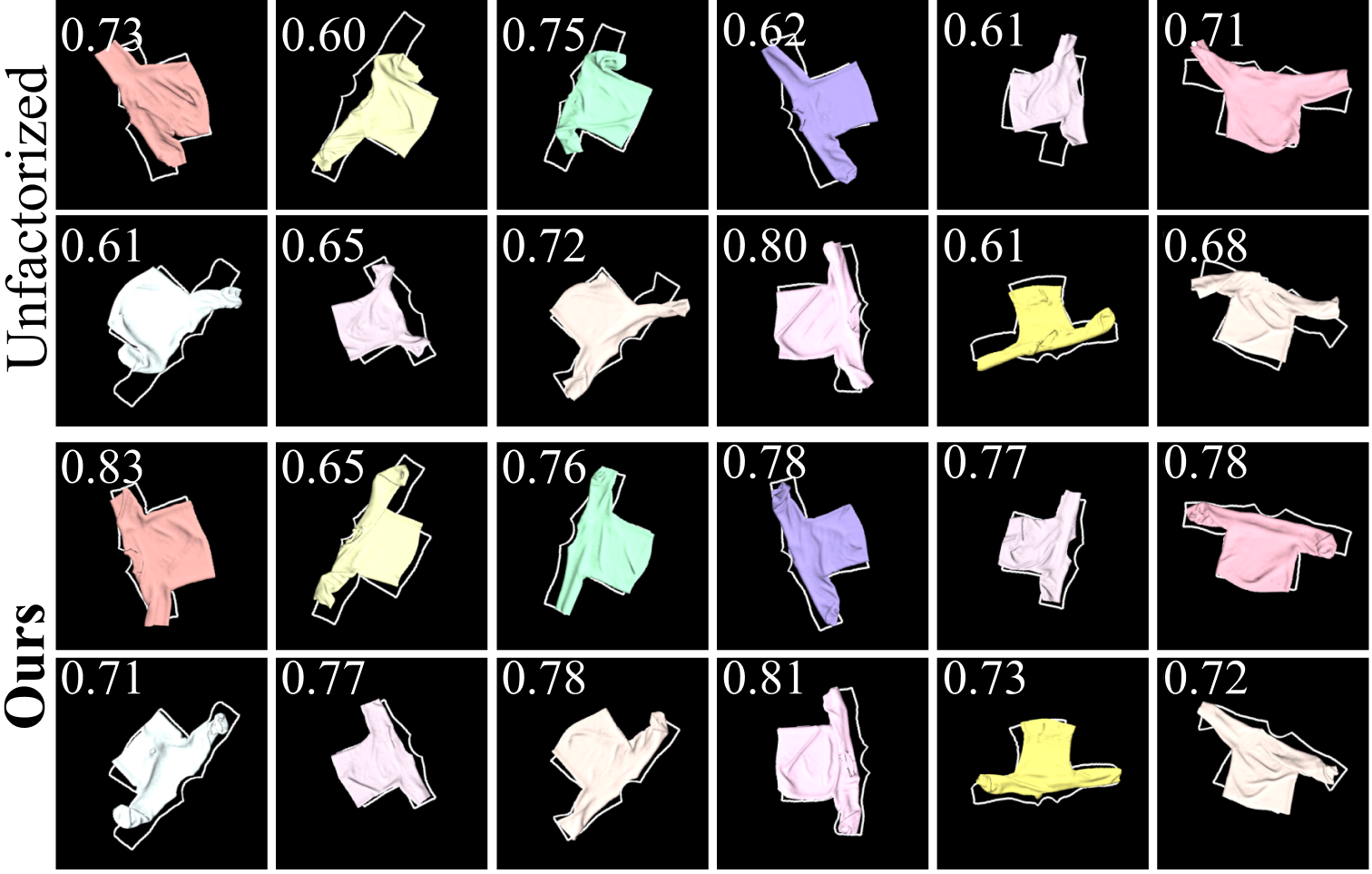}
    \vspace{-3mm}
    \caption{\textbf{Reward Comparisons. }
    We qualitatively compare the final configuration achieved by policies trained with our factorized and the unfactorized reward formulation.
    For the qualitative comparisons, the IoU of the final frame of various rollouts are shown in top-left corner of each square.
    }
    \label{fig:reward_comparison}
    \vspace{-5mm}
\end{figure}

\subsection{Metrics}
After running each policy for 10 steps, we evaluate the final 
$\rewardBaseline$~\eqref{eq:unfactorized}, $\rewardAlign$~\eqref{eq:rigid}, and $\rewardCanon$~\eqref{eq:deform} in the episode.
While these rewards account for the full configuration of the garment, they measure distance based on effectively ground-truth vertex correspondence.
This means they give larger distances for radially symmetric garments, like skirts and dresses, and they can't readily be computed on real world garments.
To address these shortcomings, we also evaluate the IoU and percent coverage from the current cloth binary image mask and the goal cloth binary image mask.
In all tables, $\downarrow$ indicates that lower is better while $\uparrow$ indicates that larger values are preferred.

\subsection{Task Generation}

Our task datasets contain randomized initial configurations of a filtered\footnote{
    Since CLOTH3D meshes are automatically generated, we manually filter unrealistic mesh examples (\eg arms as wide as cloth body), and we ensured all cloth meshes are shorter than 0.7m. 
} subset of meshes from Cloth3D~\cite{bertiche2020cloth3d}, whose configurations are generated as follows:
\begin{enumerate}
    \item \textbf{Hard Tasks} have low coverage and severe self-occlusion.
    They are generated by randomly rotating the cloth, picking a random point on the cloth, dropping it from a random height in $[0.5, 1.5]\mathrm{m}$, and then translating the cloth by a random distance in $[0.0, 0.2]\mathrm{m}$. 
    \item \textbf{Easy Tasks} have high coverage to test policies' abilities to perform small adjustments crucial to canonicalization. 
    They are generated by starting with the canonicalized configuration, and dragging a random point on the cloth by an angle uniformly sampled from $[0, 360]$ degrees by a distance uniformly sampled in $[0.5, 1]\mathrm{m}$.
\end{enumerate}
Each garment category has 2000 training and 400 testing tasks with unseen meshes, with a 75-25 and 50-50 split between hard and easy tasks respectively.

\subsection{Reward Ablation}
\label{sec:eval:reward_formulation}
We compare the canonicalized-alignment performance between the unfactorized~\eqref{eq:unfactorized} and our factorized reward formulation~\eqref{eq:factorized} on the long sleeve category.
We observe that our approach consistently does best on all metrics (Tab.~\ref{tab:reward_formulation}), 
reflected in qualitatively more consistent canonicalized-alignment (Fig.~\ref{fig:reward_comparison}).
We hypothesize that the $\rewardBaseline$ baseline struggles to canonicalize properly due to faint supervision on small deformable adjustments.
Meanwhile, our approach can emphasize canonicalization more with $\alpha = 0.6$.

\begin{table}
\vspace{-2mm}
\caption{ 
        \footnotesize
        \centering
        \textbf{Reward Ablation on Hard Tasks}
    }
    \centering
    \vspace{-2mm}
    \begin{tabular}{l|ccccc}
        \toprule
        \textbf{Metric}            &  $\mathbf{\rewardBaseline} \downarrow$   & $\mathbf{\rewardAlign}\downarrow$ &  $\mathbf{\rewardCanon} \downarrow$ & \textbf{IoU} $\uparrow$  & \textbf{Cov.}$\uparrow$ \\
        \midrule
         $\mathbf{\rewardBaseline}$       & 0.093          & 0.069          & 0.064               & 0.684          & 0.879             \\
         $\mathbf{\rewardOurs}$  ($\alpha=0.6$) & \textbf{0.075} & \textbf{0.051} & \textbf{0.052}      & \textbf{0.728} & \textbf{0.887}             \\
        \bottomrule
    \end{tabular}
    
    \vspace{-3mm}
    \label{tab:reward_formulation}
\end{table}

\subsection{Effectiveness of Combined Primitives}
\label{sec:eval:combined_primitives}

While high-velocity dynamic actions enable efficient unfolding, precise quasi-static actions are necessary for fine-grained adjustments involved in canonicalization.
We compare two single primitive systems, only Aligned-Fling (Flingbot~\cite{ha2021flingbot}'s fling with fling direction and location specified by the target alignment) and only Pick \& Place (P\&P), with our combined primitive system on canonicalized-alignment of long sleeve shirts.

\vspace{-3mm}
\begin{table}[H]
    \setlength{\tabcolsep}{0.12cm}
    \caption{ \footnotesize\centering\textbf{Primitive Ablation in Simulation}
    }
    \footnotesize
    \centering
    \begin{tabular}{ll|ccccc}
        \toprule
        Task & \textbf{Primitives} & $\mathbf{\rewardBaseline} \downarrow$ & $\mathbf{\rewardAlign}\downarrow$ & $\mathbf{\rewardCanon} \downarrow$ & \textbf{IoU} $\uparrow$ & \textbf{Cov.}$\uparrow$ \\
        \midrule
             & Aligned-Fling
             & 0.100               & 0.058                                 & 0.079                                   & 0.649                                     & 0.821                                             \\
        Easy & P\&P
             & 0.077               & 0.068                                 & \textbf{0.037}                          & \textbf{0.734}                            & \textbf{0.928}                                    \\
             & Aligned-Fling+P\&P
             & \textbf{0.075}      & \textbf{0.051}                        & 0.044                                   & 0.731                                     & 0.924                                             \\
        \midrule
             & Aligned-Fling       & 0.100                                 & 0.058                                   & 0.079                                     & 0.644                   & 0.812
        \\
        Hard & P\&P                & 0.136                                 & 0.111                                   & 0.077                                     & 0.601                   & 0.792
        \\
             & Aligned-Fling+P\&P  & \textbf{0.075}                        & \textbf{0.051}                          & \textbf{0.052}                            & \textbf{0.728}          & \textbf{0.887}
        \\
        \bottomrule
    \end{tabular}

    \vspace{-3mm}
    \label{tab:primitive:sim:easy}
\end{table}

On easy tasks in simulation (Tab.~\ref{tab:primitive:sim:easy}, top), P\&P and our combined primitive system perform similarly, confirming that quasi-static actions alone are effective for small adjustments.
In contrast, the fling-only system performs poorly because the imprecise flinging actions are ill-suited for cases where fine-grained manipulation is required (\eg Fig.~\ref{fig:pipeline}).

For hard tasks (Tab.~\ref{tab:primitive:sim:easy} bottom in simulation, Tab.~\ref{tab:primitive:real} in real), we observe that both single primitive baselines perform poorly.
Notably, Fling achieves a similar performance between easy and hard tasks in simulation, suggesting that the effect of coarse grain high-velocity actions can be effective for unfolding but is not versatile enough for canonicalization.
Meanwhile, the P\&P baseline performed significantly worse on hard tasks in simulation, confirming the findings reported in FlingBot~\cite{ha2021flingbot}.
Our combined primitive system achieves the best performance, demonstrating the synergy between coarse-grain high-velocity flings and fine-grain quasi-static actions for canonicalized-alignment.

\begin{figure}
    \centering
    \includegraphics[width=0.98\linewidth]{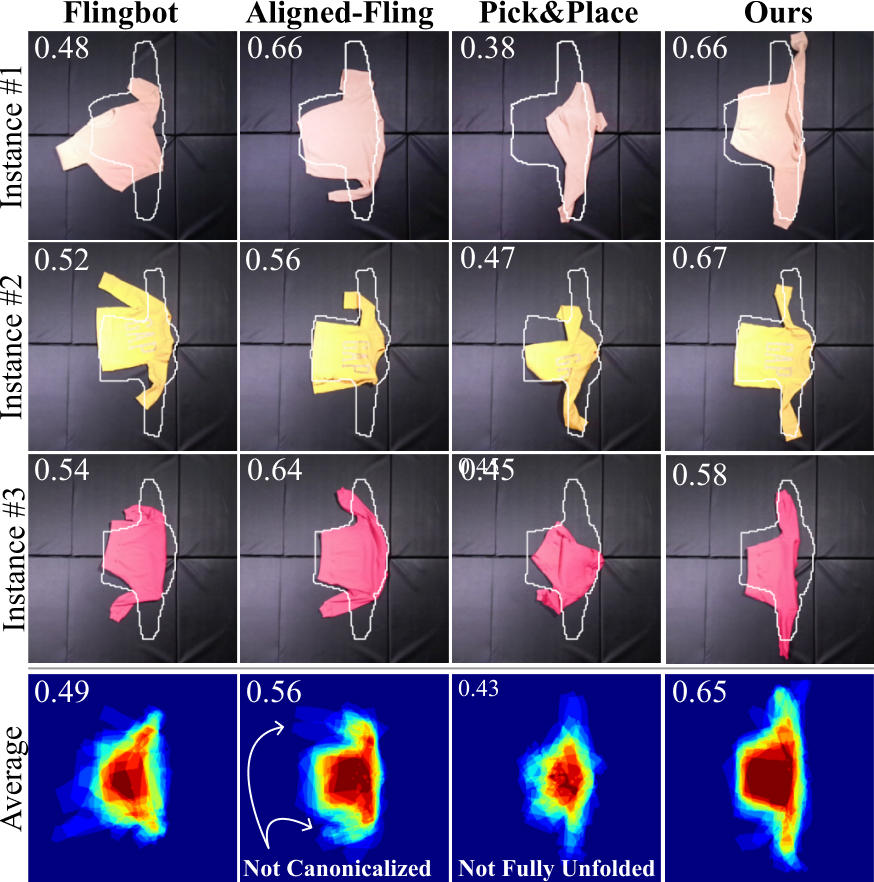} \vspace{-3mm}
    \caption{
        \textbf{Real-world Canonicalized-Alignment.}
        High-coverage configurations achieved by Flingbot aren't always aligned, which is improved with our Aligned-Fling.
        However, using only coarse-grained Aligned-Fling fails to fully canonicalize the shirt, and using only fine-grained Pick \& Place fails to fully-unfold the shirt as can be seen by the average of the final cloth masks (bottom row).
    }
    \label{fig:canal:real}  \vspace{-3mm}
\end{figure}

Due to imperfect cloth simulators, deformable dynamics such as arms twisting (instance \#2, last column in Fig.~\ref{fig:canal:real}) and stretching (instance \#3, last column in Fig.~\ref{fig:canal:real}) are never observed in simulation.
Despite this sim2real gap, our real world qualitative results (Fig.~\ref{fig:canal:real}) confirmed our simulation findings.
Specifically, we found that on average, our combined primitive approach achieved an IoU of 0.65, while other single primitive baselines/ablations reached 0.56 or less.

\subsection{Canonicalized-Alignment on Garment Category}
\label{sec:eval:multicategory}

Using the same learning and system configuration, we trained new models on 4 other garment categories, one model per garment category.
From the quantitative evaluation in Tab.~\ref{tab:multicategory}, our approach achieves around 70 IoU for all categories, which demonstrates the generality of our problem formulation with respect to garment categories.
Further, from Fig.~\ref{fig:canal:sim}, our policy has learned that while flinging is crucial for quickly unfolding crumpled garments, a few pick\&places are usually required to achieve good canonicalization.

\vspace{-2mm}
\begin{table}[H]
    \centering
    \caption{ \footnotesize
        \centering
        \textbf{Evaluation on Multiple Categories on Hard Tasks}
    }
    \vspace{-2mm}
    \begin{tabular}{l|ccccc}
        \toprule
        \textbf{Category}   & $\mathbf{\rewardBaseline} \downarrow$ & $\mathbf{\rewardAlign}\downarrow$ & $\mathbf{\rewardCanon} \downarrow$ & \textbf{IoU} $\uparrow$ & \textbf{Cov.}$\uparrow$ \\
        \midrule
        Shirt             & 0.075       & 0.051          & 0.052               & 0.728        & 0.887             \\
        Pants             & 0.098       & 0.077          & 0.053               & 0.708        & 0.892             \\
        Skirt             & 0.159       & 0.128          & 0.122               & 0.680        & 0.837             \\
        Dress             & 0.149       & 0.106          & 0.100               & 0.714        & 0.878             \\
        Jumpsuit          & 0.099       & 0.072          & 0.060               & 0.648        & 0.817             \\
        \bottomrule
    \end{tabular}
    \vspace{-3mm}
    \label{tab:multicategory}
\end{table}


\subsection{Application in Downstream Cloth Manipulation}
\label{sec:eval:downstream}
A primary motivation for this work is the improvement of downstream tasks; we hypothesize that effective canonicalized-alignment will significantly reduce the complexity of subsequent manipulation skills.
To this end, we study two garment manipulation tasks, ironing and folding (Sec.~\ref{sec:method:heuristics}).
For folding, we measured the final $\rewardBaseline$ achieved by each approach when the goal configuration is the ground-truth folded configuration at a specified alignment.
We also include a folding success rate which is a thresholded $\rewardBaseline$, where the boundary is chosen by qualitatively.



\begin{figure}
    \centering
    \includegraphics[width=0.98\linewidth]{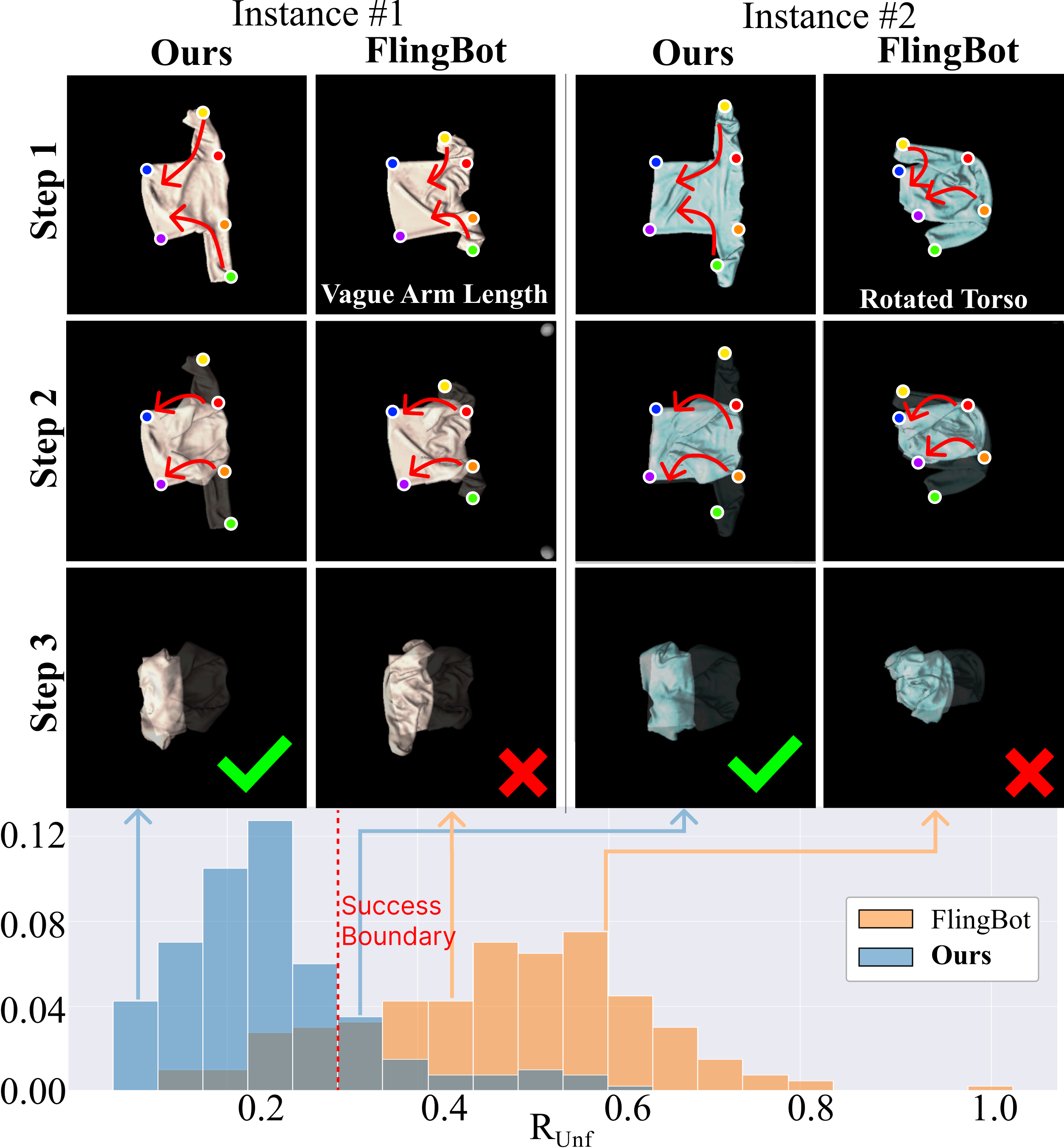}\vspace{-3mm}
    \caption{\textbf{Simulation Folding Qualitative Comparison.}  We compare our folding heuristic on our method's canonicalization results vs. FlingBot's unfolding results. We qualitatively (images) and quantitatively (histogram) demonstrate that not all high-coverage configurations ensure folding success.}
    \label{fig:downstream:folding}
\end{figure}

\textbf{Canonicalized-Alignment Improves Downstream Tasks.}
From Tab.~\ref{tab:folding_sim}, we observe that manually-designed folding heuristic completely fails at the task if starting from random initial configurations.
While success rate improves with FlingBot~\cite{ha2021flingbot}'s unfolded configurations, achieving high-coverage configurations (the goal of unfolding) does not always give structured, high-visibility configurations required by heuristics (Fig.~\ref{fig:downstream:folding}), so FlingBot~\cite{ha2021flingbot} still performs poorly.
With canonicalized-alignment, policies trained with $\rewardBaseline$ and $\rewardOurs$ achieve success rates of 84.9 and 87.8 respectively, demonstrating the importance of canonicalized-alignment reducing downstream task complexity, such that even simple manually designed task heuristics can work well.

\vspace{-3mm}
\begin{table}[H]
    \centering

    \caption{ \footnotesize
        \centering
        \textbf{Real-world Canonicalized-Alignment}
    }
    \vspace{-3mm}
    \begin{tabular}{l|cc}
        \toprule
        \textbf{Approach}              & \textbf{IoU} $\uparrow$ & \textbf{Cov.}$\uparrow$ \\
        \midrule
        FlingBot \cite{ha2021flingbot} & 0.489                      & 0.709                      \\
        Aligned-Fling                  & 0.558                      & 0.735                      \\
        P\&P                           & 0.433                      & 0.507                      \\
        Aligned-Fling+P\&P             & \textbf{0.648}                      & \textbf{0.806}                      \\
        \bottomrule
    \end{tabular}
    \label{tab:primitive:real}
\end{table}

\vspace{-3mm}
\begin{table}[H]
    \centering
    \caption{ \footnotesize
        \textbf{Simulation Folding}
    }
    \begin{tabular}{l|rr}
        \toprule
        \textbf{Approach}            & \textbf{Success} $\uparrow$ & $\mathbf{\rewardBaseline} \downarrow$ \\
        \midrule
        Random                         & 2.1                           & 0.520                                   \\
        FlingBot~\cite{ha2021flingbot} & 19.6                           & 0.486                                  \\
        $\rewardBaseline$              & 84.9                          & \textbf{0.246}                                 \\
        $\rewardOurs$ (Ours)                           & \textbf{87.8}                           & 0.253                                  \\
        \bottomrule
    \end{tabular}
    \vspace{-3mm}
    \label{tab:folding_sim}
\end{table}

\label{sec:result}

\section{Conclusion}
\label{sec:conclusion}

We introduce the task of canonicalized-alignment, a universal first step for multi-purpose garment manipulation pipelines.
By funneling a diverse array of crumpled clothing into a small set of high-visibility configurations, this task addresses much of the complexity associated with cloth's infinite DOF state space and severe self-occlusion and thus simplifies downstream tasks. After training in simulation with our novel factorized reward formulation for canonicalized alignment our learned policy generalizes to the real-world robot system and can be directly used for garment ironing and folding.
Due to imperfect cloth simulators, we hypothesize that canonicalized-alignment performance can be improved if a real-world supervision signal could be derived to enable real-world finetuning, and believe this is an interesting direction for future work.

\clearpage
\mypara{Acknowledgements.}
This work was supported in part by the Toyota Research Institute, NSF Award \#2143601, \#2037101, and \#2132519. We would like to thank Google for the UR5 robot hardware. The views and conclusions contained herein are those of the authors and should not be interpreted as necessarily representing the official policies, either expressed or implied, of the sponsors.

\bibliographystyle{unsrtnat}
\bibliography{references} 

\begin{thebibliography}{34}
\providecommand{\natexlab}[1]{#1}
\providecommand{\url}[1]{\texttt{#1}}
\expandafter\ifx\csname urlstyle\endcsname\relax
  \providecommand{\doi}[1]{doi: #1}\else
  \providecommand{\doi}{doi: \begingroup \urlstyle{rm}\Url}\fi

\bibitem[Cusumano-Towner et~al.(2011)Cusumano-Towner, Singh, Miller, O'Brien,
  and Abbeel]{cusumano2011bringing}
Marco Cusumano-Towner, Arjun Singh, Stephen Miller, James~F O'Brien, and Pieter
  Abbeel.
\newblock Bringing clothing into desired configurations with limited
  perception.
\newblock In \emph{2011 IEEE international conference on robotics and
  automation}, pages 3893--3900. IEEE, 2011.

\bibitem[Maitin-Shepard et~al.(2010)Maitin-Shepard, Cusumano-Towner, Lei, and
  Abbeel]{maitin2010cloth}
Jeremy Maitin-Shepard, Marco Cusumano-Towner, Jinna Lei, and Pieter Abbeel.
\newblock Cloth grasp point detection based on multiple-view geometric cues
  with application to robotic towel folding.
\newblock In \emph{2010 IEEE International Conference on Robotics and
  Automation}, pages 2308--2315. IEEE, 2010.

\bibitem[Triantafyllou et~al.(2016)Triantafyllou, Mariolis, Kargakos,
  Malassiotis, and Aspragathos]{triantafyllou2016geometric}
Dimitra Triantafyllou, Ioannis Mariolis, Andreas Kargakos, Sotiris Malassiotis,
  and Nikos Aspragathos.
\newblock A geometric approach to robotic unfolding of garments.
\newblock \emph{Robotics and Autonomous Systems}, 75:\penalty0 233--243, 2016.

\bibitem[Osawa et~al.(2007)Osawa, Seki, and Kamiya]{osawa2007unfolding}
Fumiaki Osawa, Hiroaki Seki, and Yoshitsugu Kamiya.
\newblock Unfolding of massive laundry and classification types by dual
  manipulator.
\newblock \emph{Journal of Advanced Computational Intelligence and Intelligent
  Informatics}, 11\penalty0 (5):\penalty0 457--463, 2007.

\bibitem[Sun et~al.(2013)Sun, Aragon-Camarasa, Cockshott, Rogers, and
  Siebert]{Sun2013AHA}
Li~Sun, G.~Aragon-Camarasa, W.~Cockshott, Simon Rogers, and J.~Siebert.
\newblock A heuristic-based approach for flattening wrinkled clothes.
\newblock In \emph{TAROS}, 2013.

\bibitem[{Willimon} et~al.(2011){Willimon}, {Birchfield}, and
  {Walker}]{willimon2011model}
B.~{Willimon}, S.~{Birchfield}, and I.~{Walker}.
\newblock Model for unfolding laundry using interactive perception.
\newblock In \emph{2011 IEEE/RSJ International Conference on Intelligent Robots
  and Systems}, pages 4871--4876, 2011.
\newblock \doi{10.1109/IROS.2011.6095066}.

\bibitem[Zhou et~al.(2021)Zhou, Zahra, Duan, Huo, Wu, and
  Navarro-Alarcon]{zhou2021folding}
Peng Zhou, Omar Zahra, Anqing Duan, Shengzeng Huo, Zeyu Wu, and David
  Navarro-Alarcon.
\newblock Learning cloth folding tasks with refined flow based spatio-temporal
  graphs.
\newblock \emph{arXiv preprint arXiv:2110.08620}, 2021.

\bibitem[Osawa et~al.(2006)Osawa, Seki, and Kamiya]{osawa2006clothes}
Fumiaki Osawa, Hiroaki Seki, and Yoshitsugu Kamiya.
\newblock Clothes folding task by tool-using robot.
\newblock \emph{Journal of Robotics and Mechatronics}, 18\penalty0
  (5):\penalty0 618--625, 2006.

\bibitem[Miller et~al.(2011)Miller, Fritz, Darrell, and
  Abbeel]{miller2011parametrized}
Stephen Miller, Mario Fritz, Trevor Darrell, and Pieter Abbeel.
\newblock Parametrized shape models for clothing.
\newblock In \emph{2011 IEEE International Conference on Robotics and
  Automation}, pages 4861--4868. IEEE, 2011.

\bibitem[Avigal et~al.(2022)Avigal, Berscheid, Asfour, Kröger, and
  Goldberg]{ken2022speedfolding}
Yahav Avigal, Lars Berscheid, Tamim Asfour, Torsten Kröger, and Ken Goldberg.
\newblock Speedfolding: Learning efficient bimanual folding of garments.
\newblock In \emph{2022 IEEE/RSJ International Conference on Intelligent Robots
  and Systems (IROS)}. IEEE, 2022.

\bibitem[Ha and Song(2021)]{ha2021flingbot}
Huy Ha and Shuran Song.
\newblock Flingbot: The unreasonable effectiveness of dynamic manipulation for
  cloth unfolding.
\newblock In \emph{Conference on Robotic Learning (CoRL)}, 2021.

\bibitem[Lee et~al.(2020)Lee, Ward, Cosgun, Dasagi, Corke, and
  Leitner]{lee2020learning}
Robert Lee, Daniel Ward, Akansel Cosgun, Vibhavari Dasagi, Peter Corke, and
  Jurgen Leitner.
\newblock Learning arbitrary-goal fabric folding with one hour of real robot
  experience, 2020.

\bibitem[Li et~al.(2019)Li, Wu, Tedrake, Tenenbaum, and
  Torralba]{li2019learning}
Yunzhu Li, Jiajun Wu, Russ Tedrake, Joshua~B Tenenbaum, and Antonio Torralba.
\newblock Learning particle dynamics for manipulating rigid bodies, deformable
  objects, and fluids.
\newblock In \emph{ICLR}, 2019.

\bibitem[Bersch et~al.(2011)Bersch, Pitzer, and Kammel]{bersch2011bimanual}
Christian Bersch, Benjamin Pitzer, and S{\"o}ren Kammel.
\newblock Bimanual robotic cloth manipulation for laundry folding.
\newblock In \emph{2011 IEEE/RSJ International Conference on Intelligent Robots
  and Systems}, pages 1413--1419. IEEE, 2011.

\bibitem[Doumanoglou et~al.(2016)Doumanoglou, Stria, Peleka, Mariolis, Petrik,
  Kargakos, Wagner, Hlav{\'a}{\v{c}}, Kim, and
  Malassiotis]{doumanoglou2016folding}
Andreas Doumanoglou, Jan Stria, Georgia Peleka, Ioannis Mariolis, Vladimir
  Petrik, Andreas Kargakos, Libor Wagner, V{\'a}clav Hlav{\'a}{\v{c}}, Tae-Kyun
  Kim, and Sotiris Malassiotis.
\newblock Folding clothes autonomously: A complete pipeline.
\newblock \emph{IEEE Transactions on Robotics}, 32\penalty0 (6):\penalty0
  1461--1478, 2016.

\bibitem[Tanaka et~al.(2007)Tanaka, Kamotani, and Yokokohji]{tanaka2007origami}
Kenta Tanaka, Yusuke Kamotani, and Yasuyoshi Yokokohji.
\newblock Origami folding by a robotic hand.
\newblock In \emph{2007 IEEE/RSJ International Conference on Intelligent Robots
  and Systems}, pages 2540--2547. IEEE, 2007.

\bibitem[Balkcom and Mason(2008)]{balkcom2008robotic}
Devin~J Balkcom and Matthew~T Mason.
\newblock Robotic origami folding.
\newblock \emph{The International Journal of Robotics Research}, 27\penalty0
  (5):\penalty0 613--627, 2008.

\bibitem[Stria et~al.(2014)Stria, Průša, Hlaváč, Wagner, Petrík, Krsek,
  and Smutný]{stria2014garment}
Jan Stria, Daniel Průša, Václav Hlaváč, Libor Wagner, Vladimír Petrík,
  Pavel Krsek, and Vladimír Smutný.
\newblock Garment perception and its folding using a dual-arm robot.
\newblock In \emph{2014 IEEE/RSJ International Conference on Intelligent Robots
  and Systems}, pages 61--67, 2014.
\newblock \doi{10.1109/IROS.2014.6942541}.

\bibitem[Xu et~al.(2022)Xu, Chi, Burchfiel, Cousineau, Feng, and
  Song]{xu2022dextairity}
Zhenjia Xu, Cheng Chi, Benjamin Burchfiel, Eric Cousineau, Siyuan Feng, and
  Shuran Song.
\newblock Dextairity: Deformable manipulation can be a breeze.
\newblock In \emph{Proceedings of Robotics: Science and Systems (RSS)}, 2022.

\bibitem[Huang et~al.(2022)Huang, Lin, and Held]{huang2022mesh}
Zixuan Huang, Xingyu Lin, and David Held.
\newblock Mesh-based dynamics with occlusion reasoning for cloth manipulation.
\newblock \emph{arXiv preprint arXiv:2206.02881}, 2022.

\bibitem[Lin et~al.(2022)Lin, Wang, Huang, and Held]{lin2022learning}
Xingyu Lin, Yufei Wang, Zixuan Huang, and David Held.
\newblock Learning visible connectivity dynamics for cloth smoothing.
\newblock In \emph{Conference on Robot Learning}, pages 256--266. PMLR, 2022.

\bibitem[Matas et~al.(2018)Matas, James, and Davison]{matas2018sim}
Jan Matas, Stephen James, and Andrew~J Davison.
\newblock Sim-to-real reinforcement learning for deformable object
  manipulation.
\newblock In \emph{Conference on Robot Learning}, pages 734--743. PMLR, 2018.

\bibitem[Hietala et~al.(2021)Hietala, Blanco-Mulero, Alcan, and
  Kyrki]{hietala2021closing}
Julius Hietala, David Blanco-Mulero, Gokhan Alcan, and Ville Kyrki.
\newblock Closing the sim2real gap in dynamic cloth manipulation.
\newblock \emph{arXiv preprint arXiv:2109.04771}, 2021.

\bibitem[Jangir et~al.(2020)Jangir, Alenya, and Torras]{jangir2020dynamic}
Rishabh Jangir, Guillem Alenya, and Carme Torras.
\newblock Dynamic cloth manipulation with deep reinforcement learning.
\newblock In \emph{2020 IEEE International Conference on Robotics and
  Automation (ICRA)}, pages 4630--4636. IEEE, 2020.

\bibitem[Tsurumine et~al.(2019)Tsurumine, Cui, Uchibe, and
  Matsubara]{tsurumine2019deep}
Yoshihisa Tsurumine, Yunduan Cui, Eiji Uchibe, and Takamitsu Matsubara.
\newblock Deep reinforcement learning with smooth policy update: Application to
  robotic cloth manipulation.
\newblock \emph{Robotics and Autonomous Systems}, 112:\penalty0 72--83, 2019.

\bibitem[Seita et~al.(2020)Seita, Ganapathi, Hoque, Hwang, Cen, Tanwani,
  Balakrishna, Thananjeyan, Ichnowski, Jamali, et~al.]{seita2020deep}
Daniel Seita, Aditya Ganapathi, Ryan Hoque, Minho Hwang, Edward Cen, Ajay~Kumar
  Tanwani, Ashwin Balakrishna, Brijen Thananjeyan, Jeffrey Ichnowski, Nawid
  Jamali, et~al.
\newblock Deep imitation learning of sequential fabric smoothing from an
  algorithmic supervisor.
\newblock In \emph{2020 IEEE/RSJ International Conference on Intelligent Robots
  and Systems (IROS)}, pages 9651--9658. IEEE, 2020.

\bibitem[Weng et~al.(2022)Weng, Bajracharya, Wang, Agrawal, and
  Held]{weng2022fabricflownet}
Thomas Weng, Sujay~Man Bajracharya, Yufei Wang, Khush Agrawal, and David Held.
\newblock Fabricflownet: Bimanual cloth manipulation with a flow-based policy.
\newblock In \emph{Conference on Robot Learning}, pages 192--202. PMLR, 2022.

\bibitem[Tanaka et~al.(2018)Tanaka, Arnold, and Yamazaki]{tanaka2018emd}
Daisuke Tanaka, Solvi Arnold, and Kimitoshi Yamazaki.
\newblock Emd net: An encode--manipulate--decode network for cloth
  manipulation.
\newblock \emph{IEEE Robotics and Automation Letters}, 3\penalty0 (3):\penalty0
  1771--1778, 2018.

\bibitem[LeCun et~al.(2010)LeCun, Kavukcuoglu, and
  Farabet]{lecun2010convolutional}
Yann LeCun, Koray Kavukcuoglu, and Cl{\'e}ment Farabet.
\newblock Convolutional networks and applications in vision.
\newblock In \emph{Proceedings of 2010 IEEE international symposium on circuits
  and systems}, pages 253--256. IEEE, 2010.

\bibitem[Wu et~al.(2020{\natexlab{a}})Wu, Sun, Zeng, Song, Lee, Rusinkiewicz,
  and Funkhouser]{Wu_2020}
Jimmy Wu, Xingyuan Sun, Andy Zeng, Shuran Song, Johnny Lee, Szymon
  Rusinkiewicz, and Thomas Funkhouser.
\newblock Spatial action maps for mobile manipulation.
\newblock In \emph{Proceedings of Robotics: Science and Systems (RSS)},
  2020{\natexlab{a}}.

\bibitem[Wu et~al.(2020{\natexlab{b}})Wu, Sun, Zeng, Song, Lee, Rusinkiewicz,
  and Funkhouser]{wu2020spatial}
Jimmy Wu, Xingyuan Sun, Andy Zeng, Shuran Song, Johnny Lee, Szymon
  Rusinkiewicz, and Thomas Funkhouser.
\newblock Spatial action maps for mobile manipulation.
\newblock In \emph{Proceedings of Robotics: Science and Systems (RSS)},
  2020{\natexlab{b}}.

\bibitem[Kingma and Ba(2014)]{kingma2014adam}
Diederik~P Kingma and Jimmy Ba.
\newblock Adam: A method for stochastic optimization.
\newblock \emph{arXiv preprint arXiv:1412.6980}, 2014.

\bibitem[Chen et~al.(2019)Chen, Papandreou, Schroff, and
  Adam]{chen2019rethinking}
Liang-Chieh Chen, George Papandreou, Florian Schroff, and Hartwig Adam.
\newblock Rethinking atrous convolution for semantic image segmentation. arxiv
  2017.
\newblock \emph{arXiv preprint arXiv:1706.05587}, 2, 2019.

\bibitem[Bertiche et~al.(2020)Bertiche, Madadi, and
  Escalera]{bertiche2020cloth3d}
Hugo Bertiche, Meysam Madadi, and Sergio Escalera.
\newblock Cloth3d: Clothed 3d humans.
\newblock In \emph{European Conference on Computer Vision}, pages 344--359.
  Springer, 2020.

\end{thebibliography}

\end{document}